\documentclass[runningheads]{llncs}

\usepackage{makeidx}  
\usepackage{url}
\usepackage{color}
\usepackage{longtable}
\usepackage{multirow}
\usepackage{graphicx}
\usepackage{amssymb, amsmath, bm}
\usepackage{mathtools}
\usepackage{mathrsfs}
\usepackage{booktabs}
\usepackage{cite}
\usepackage{dsfont}
\usepackage[colorlinks,linkcolor=blue]{hyperref}
\usepackage{makecell}
\usepackage[noend]{algpseudocode}
\usepackage{algorithmicx,algorithm}
\usepackage{graphicx}

\usepackage[misc]{ifsym} 
\usepackage{bbding}
\begin{document}
	
	\title{Flip Learning: Erase to Segment}
	\titlerunning{Flip Learning: Erase to Segment}

	\author{Yuhao Huang\inst{1,2,3}\thanks{Yuhao Huang and Xin Yang contribute equally to this work.}, Xin Yang\inst{1,2,3\star} \and Yuxin Zou\inst{1,2,3} \and Chaoyu Chen\inst{1,2,3} \and \\Jian Wang\inst{1,2,3} \and Haoran Dou\inst{4} \and Nishant Ravikumar\inst{4,5} \and Alejandro F Frangi\inst{1,4,5,6} \and Jianqiao Zhou\inst{7} \and Dong Ni\inst{1,2,3}\textsuperscript{(\Letter)}} 
	
	
	\authorrunning{Huang et al.}
	\institute{
		\textsuperscript{$1$}National-Regional Key Technology Engineering Laboratory for Medical Ultrasound, School of Biomedical Engineering, Health Science Center, Shenzhen University, China\\
		\email{nidong@szu.edu.cn} \\
		\textsuperscript{$2$}Medical Ultrasound Image Computing (MUSIC) Lab, Shenzhen University, China\\
		\textsuperscript{$3$}Marshall Laboratory of Biomedical Engineering, Shenzhen University, China\\
		\textsuperscript{$4$}Centre for Computational Imaging and Simulation Technologies in Biomedicine (CISTIB), University of Leeds, UK\\ 
		\textsuperscript{$5$}Leeds Institute of Cardiovascular and Metabolic Medicine, University of Leeds, UK\\
		\textsuperscript{$6$}Medical Imaging Research Center (MIRC), KU Leuven, Leuven, Belgium\\
		\textsuperscript{$7$}Department of Ultrasound Medicine, Ruijin Hospital, School of Medicine, Shanghai Jiaotong University, China\\ }
	\maketitle              
	\begin{abstract}
	Nodule segmentation from breast ultrasound images is challenging yet essential for the diagnosis.
	Weakly-supervised segmentation (WSS) can help reduce time-consuming and cumbersome manual annotation. 
	Unlike existing weakly-supervised approaches, in this study, we propose a novel and general WSS framework called Flip Learning, which only needs the box annotation. 
	Specifically, the target in the label box will be erased gradually to flip the classification tag, and the erased region will be considered as the segmentation result finally. 
	Our contribution is three-fold. 
	First, our proposed approach erases on superpixel level using a Multi-agent Reinforcement Learning framework to exploit the prior boundary knowledge and accelerate the learning process.
	Second, we design two rewards: classification score and intensity distribution reward, to avoid under- and over-segmentation, respectively.
	Third, we adopt a coarse-to-fine learning strategy to reduce the residual errors and improve the segmentation performance. 
	Extensively validated on a large dataset, our proposed approach achieves competitive performance and shows great potential to narrow the gap between fully-supervised and weakly-supervised learning.
	\keywords{Ultrasound \and Weakly-supervised segmentation\and Reinforcement learning}
	\end{abstract}
	
	\section{Introduction}
	\label{sec:intro}
	Nodule segmentation in breast ultrasound (US) is important for quantitative diagnostic procedures and treatment planning. 
	Image segmentation's performance has been significantly advanced by the recent availability of fully-supervised segmentation methods~\cite{liu2019deep,minaee2020image}. 
	However, training such methods usually relies on the availability of pixel-level masks laboriously and manually annotated by sonographers.
	Hence, designing an automatic weakly-supervised segmentation (WSS) based system that only requires coarse labels, e.g. bounding box (BBox), is desirable to ease the pipeline of manual annotation and save time for sonographers.
	
\begin{figure*}[!h]
		\centering
		\includegraphics[width=0.8\linewidth]{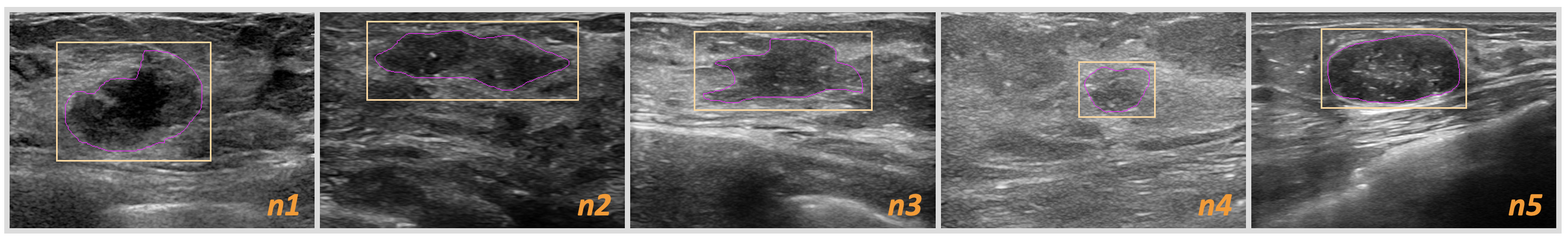}
		\caption{Breast nodule images in 2D US with annotated box and boundary.}
		\label{fig:intro}
	\end{figure*}

	Breast cancer occurs in the highest frequency in women among all cancers and is also one of the leading causes of cancer death worldwide~\cite{siegel2021cancer}. %
	Thus, extracting the nodule boundary is essential for detecting and diagnosing breast cancer at its early stage.
	As shown in Fig.~\ref{fig:intro}, segmenting the nodule's boundary from the US image with weak annotation, i.e., BBox, is still very challenging. 
	First, nodules of the same histological type may present completely different US image characteristics because of variances in their disease differentiation and stage. 
	Second, nodules of different types have extremely high inter-class differences and often display varied appearance patterns, making designing machine learning algorithms difficult.
	The third challenge is that different tissues of US images have different echo characteristics. Therefore, the intensity distribution of foreground and background in different US images also has great diversity.
	
	In the WSS literature, Class Activation Mapping (CAM) based methods~\cite{zhou2016learning, selvaraju2017grad,chattopadhay2018grad} were proposed to visualise the most discriminative features and regions of interest obtained by the classifier.
	Wei et al.~\cite{wei2017object} proposed to erase the CAM area predicted by the classifier constantly to optimise its performance on the WSS task. 
	However, classifiers are only responsive to small and sparse discriminative regions from the object of interest, which deviates from the segmentation task requirement that needs to localise dense, interior and integral regions for pixel-wise inference.
	Therefore, the above methods ignored the pairing relationship between pixels in the image and strongly relied on the CAMs with well positioning and coverage performance.
	Except for the image-level WSS approaches based on inaccurate CAMs, some methods employed box annotations to obtain high-quality prediction masks at a small annotation cost.
	However, most of them, such as BoxSup~\cite{dai2015boxsup}, SDI~\cite{khoreva2017simple}  and Box2Seg~\cite{kulharia2020box2seg}, highly rely on pseudo-mask generation algorithms (e.g., MCG~\cite{pont2016multiscale} or GrabCut~\cite{rother2004grabcut} based on object shape priors).
	Thus, they may not suit US image segmentation tasks. 
	
	In this study, we propose a novel and general Flip Learning framework for BBox based WSS.
	We believe our proposed framework is totally different from the current existing WSS methods. 
	Our contribution is three-fold.
	First, the erasing process via Multi-agent Reinforcement Learning (MARL) is based on superpixels, which can capture the prior boundary information and improve learning efficiency.
	Second, we carefully design two rewards for guiding the agents accurately. Specifically, the classification score reward (CSR) is used for pushing the agents' erasing for label flipping, while the intensity distribution reward (IDR) is employed to limit the over-segmentation.
	Third, we employ a coarse-to-fine (C2F) strategy to simplify agents' learning for residuals decreasing and segmentation performance improvement.
	Validation experiments demonstrated that the proposed Flip Learning framework could achieve high accuracy in the nodule segmentation task.

	\section{Method}
	\label{sec:method}
	\begin{figure*}[!t]
		\centering
		\includegraphics[width=0.97\linewidth]{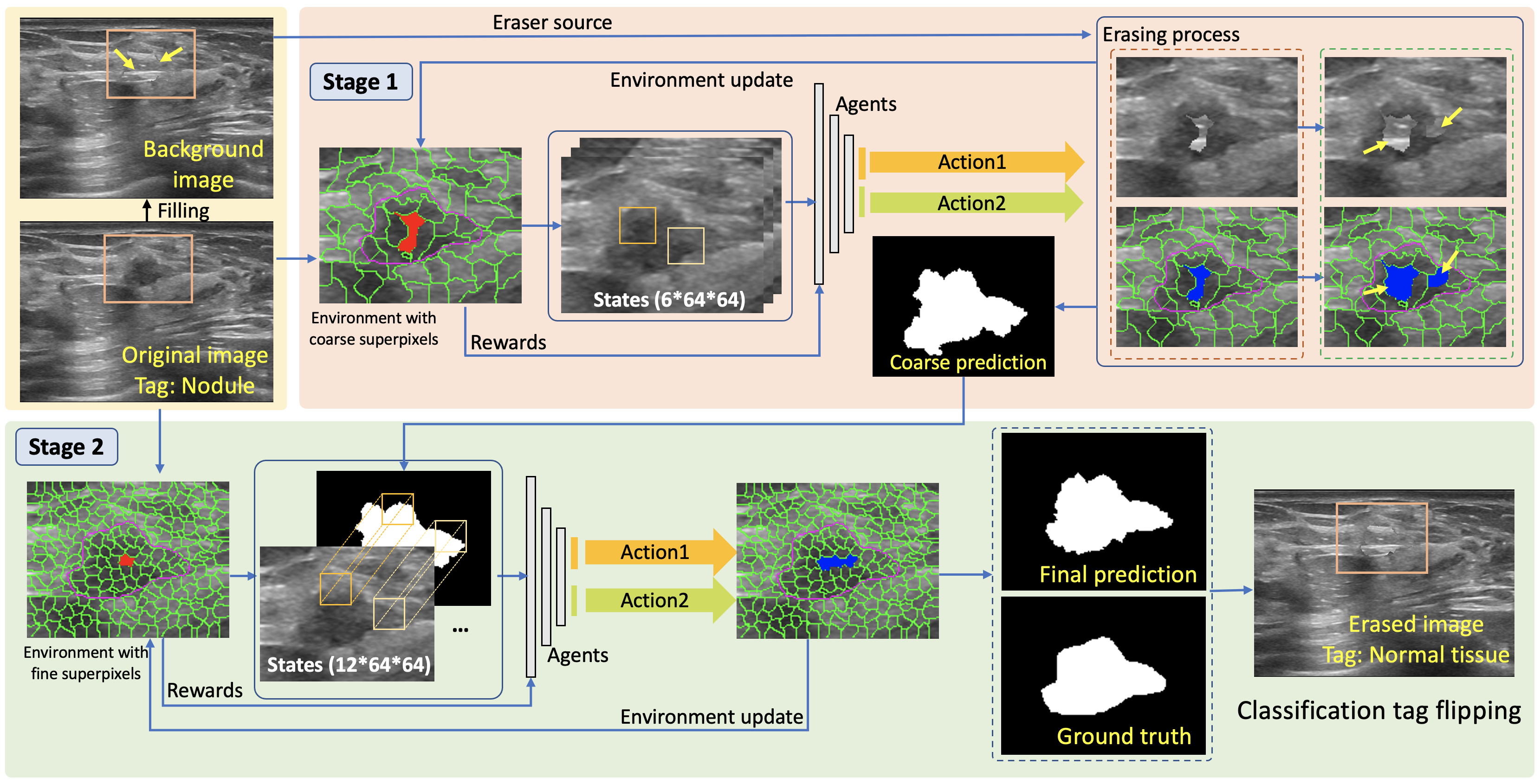}
		\caption{The workflow of our proposed framework.}
		\label{fig:framework}
	\end{figure*}

	Fig.~\ref{fig:framework} shows the workflow of our proposed framework for nodule segmentation in 2D US images. The proposed Flip Learning framework is based on MARL, in which the agents erase the nodule from its BBox. The classification score of the nodule will decrease progressively, and the tag will be flipped. The erased region will be taken as the final segmentation prediction. In our erase-to-segment system, 
	we first generate a background image to provide the eraser source to fill the erased region suitably to be indistinguishable from normal tissue. We further generate superpixels in the BBox for prior boundary extraction and thus improve learning efficiency. Then, a C2F strategy is employed to obtain an accurate segmentation result effectively.

	\subsubsection{Patch-level Copy-Paste for Precise Erasing Source.} 
\begin{figure*}[t]
		\centering
		\includegraphics[width=0.85\linewidth]{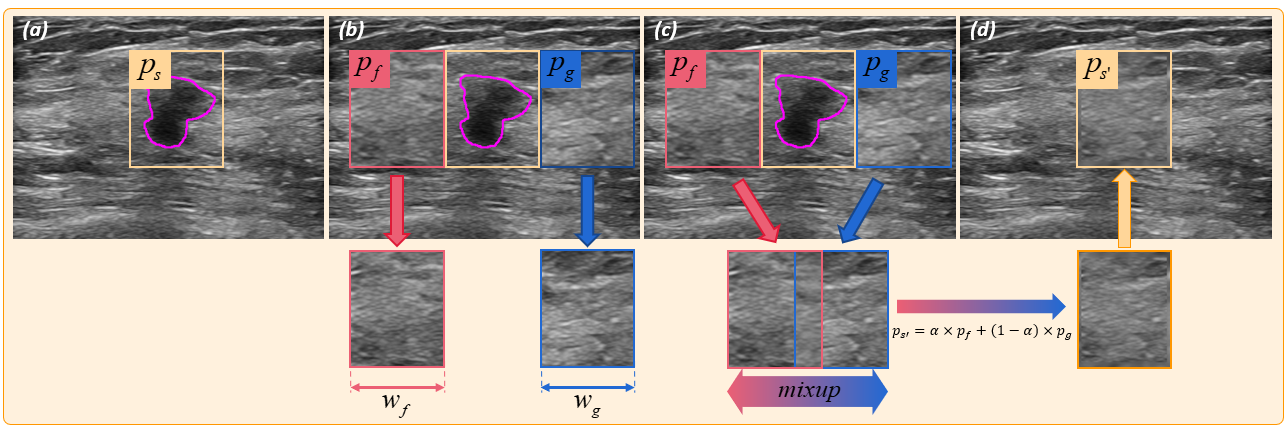}
		\caption{Example of patch-level copy-paste.}
		\label{fig:fillingg}
	\end{figure*}

US images of most normal tissues display a certain degree of local texture and gray-scale continuity.
	To trade-off between algorithm complexity and performance, we present a simple and feasible method to fill the BBox area.
	As shown in Fig.~\ref{fig:fillingg}: (a)-(b) According to the context information, we first obtain the patches ($p_{f}$ and $p_{g}$ with width $w_{f}$ and $w_{g}$, respectively) in the original image as region proposals. These patches are similar to the background around the annotated box.
	(c) Next, we adopt a $mixup$~\cite{Zhang2017mixup} to copy and paste these region proposals to generate the pseudo-background image (see Equ.~\ref{equccc}).
	(d) Finally, to achieve a higher quality fusion between the pseudo-background and the real background, we adopt a local-edge mean filter to optimise the copy-paste edge area $p_{s'}$.
	\noindent
	\begin{equation}
	\label{equccc}
		p_{s'}(x,y)=\alpha \times p_{f}(x,y) + (1-\alpha) \times p_{g}(x,y),~~~\alpha=\frac{x}{(w_{f}+w_{g})/2}
	\end{equation}

	\subsubsection{MARL Framework for Efficient and Accurate Erasing.}
	Recently, RL-based methods have shown great potential in various medical imaging tasks~\cite{yang2021searching,zhou2021deep}.
	To accelerate the learning process, we propose to use a MARL framework with two agents erasing the object simultaneously. 
	We further define the MARL erasing framework using these elements:
	
	 $\textbf{-Environment:}$ The $Environment$ is defined as the BBox area of the original US image. It contains the object to be segmented and limits the agents' moving. 
	However, it is noted that erasing based on pixel-level is difficult and inefficient due to the weak supervised signal. 
	Thus, we use superpixel algorithms for enhancing the supervised signal, which group pixels with similar properties into perceptually meaningful atomic regions while considering spatial constraints. Additionally, it preserves the edge information of the original image during enhancing the local consistency. 
	As explained above, in this study, we generate superpixels for the BBox area to obtain a superpixel-level environment before erasing. Specifically, we adopt Superpixels Extracted via EnergyDriven Sampling (SEEDS) to obtain superpixel-blocks (details refer to~\cite{2012SEEDS}).
	
	$\textbf{-Agents:}$ The $Agent$ is to learn a policy for segmentation by interacting with the superpixel-level environment. In our study, two agents $Agent_{k,k=1,2}$ share the parameters in the convolution layers for knowledge sharing, and have independent fully connected layers for decision-making. They traverse the whole BBox with each of them handling approximately half of the total superpixels, and the centre superpixel of BBox is set as start point for each agent.

	$\textbf{-States:}$ The $State$ of one agent can be defined as a 64$\times$64 area, with its centre indicates the agent's position. We further define $States$ as the last six areas observed by two agents, with the size of 6$\times$64$\times$64. States concatenating can provide rich information, which can promote agents' learning.
	
	$\textbf{-Actions:}$ The $Action$ space of our framework contains only two operations, including $erasing$ and $passing$. Note that the action is taken on the superpixel level. The action $erasing$ indicates that this superpixel will be erased, and the region will be filled according to the generated background image. The action $passing$ represents the agents choose not to erase this superpixel region.
	
\begin{figure*}[!t]
		\centering
		\includegraphics[width=0.95\linewidth]{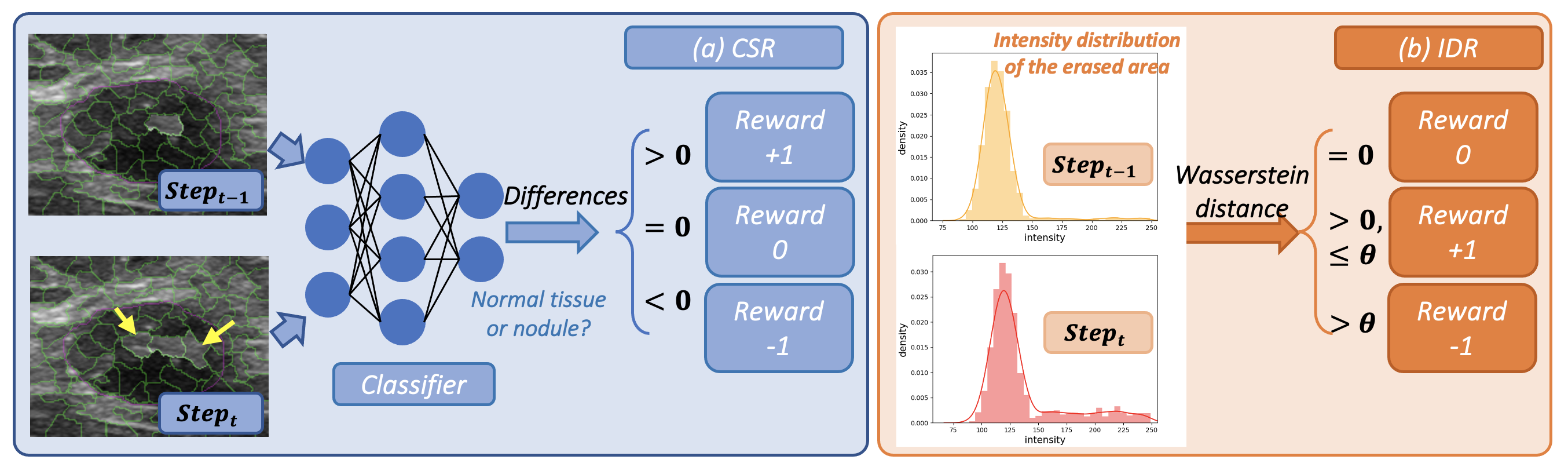}
		\caption{Details of (a) CSR and (b) IDR. Note that we first train a classifier to classify `nodule' and `normal tissue', and obtain CSR by calculating the differences of nodule scores between $step_{t-1}$ and $step_{t}$. }
		\label{fig:reward}
	\end{figure*}

	$\textbf{-Rewards:}$ The $Reward$ guides the agents' erasing process. As shown in Fig.~\ref{fig:reward}, we design two rewards, including 1) CSR and 2) IDR. Specifically, CSR is a basic reward used to guide the agents to erase the object from the BBox for classification tag flipping. However, using such reward separately may cause over-segmentation because the agents may tend to fill the whole BBox to make a high score. Thus, we introduce additional IDR to overcome this problem. Specifically, if one superpixel is erased in $step_t$ and makes the intensity distribution ($I_t$) of the erased area highly different from that in $step_{t-1}$, the agent will be punished. The difference between the two distributions is calculated using the Wasserstein distance~\cite{villani2008optimal}. The total reward $R_k$ for each $Agent_{k}$ can be defined as:
	\noindent
	\begin{equation}
		R_k= sgn(Sc_{t-1}-Sc_{t}) + thr(W(I_{t-1},I_{t}),\theta),
		\label{equ0}
	\end{equation}
	where $sgn$ and $thr$ are the sign and threshold function, respectively. $Sc_t$ represents the nodule scores in $step_t$. W($\cdot$) calculates the Wasserstein distance between the erased region's intensity distribution in $step_{t-1}$ and $step_{t}$, and $\theta=25$ is the pre-set threshold. More details can be seen in Fig.~\ref{fig:reward}. 
	
	$\textbf{-Terminal signals:}$ The terminal signal represents when to terminate the agent-environment interaction. We adopt two types of terminated strategy in our study: 1) attaining the maximum number of traversals $\mathcal{N}=2$ and 2) classification score of nodule less than a pre-set threshold $\beta=0.05$. Such a termination strategy can balance efficiency and accuracy during both training and testing.

	In deep Q-network (DQN)~\cite{mnih2015human}, both selecting and evaluating an action use a $max$ operation, which may cause over-estimation for Q-values and lead to unstable training. Double DQN (DDQN)~\cite{van2016deep} is then proposed to decouple the action selection from evaluation to stabilise the learning process. In this study, for improving learning efficiency, we adopt DDQN with a naive prioritised replay buffer~\cite{schaul2016prioritized} with the size of $\mathcal{M}$, which contains sequences of state $s$, action $a$, reward $r$ and next state $s'$. The $q_{target}$ can be calculated by:
	\begin{equation}
		q_{target} = r + \gamma Q(s',\mathop {argmax}\limits_{a'}Q(s',a';\omega);\omega'),
		\label{q_target}
	\end{equation}
	Q($\omega$) and Q($\omega'$) represent current and target Q-networks, respectively. $\gamma$ is the discount factor that balances the importance of current and future rewards. Then, with the uniform sampling $U(\cdot)$, and the loss function can be written by:
	\begin{equation}
		L = \mathbb{E}_{s,r,a,s' \sim U(\mathcal{M})}(q_{target}-Q(s,a;\omega))^{2}
		\label{loss}
	\end{equation}
	
	
	\subsubsection{C2F Learning Strategy.} 
	
	Finer superpixel implies more accurate boundary prior knowledge, leading to a higher upper bound of segmentation performance. However, learning on the BBox with fine-superpixel directly may be too difficult for the agents. The agents may get into local optimality easily due to the weak supervised signal. Thus, the segmentation result may contain residual errors. Inspired by~\cite{tu2008auto}, we adopt a two-stage strategy for reducing the residual errors and improving the segmentation performance. Specifically, in the first stage of learning, the agents learn on a coarse superpixel BBox and output a coarse prediction (see $Stage~1$ in Fig.~\ref{fig:framework}). Such coarse predictions are leveraged to make a channel-wise concatenation with original images as the second stage's input, which can provide additional information and help the agent learn on the BBox with fine-superpixel efficiently (see $Stage~2$ in Fig.~\ref{fig:framework}). 

	\section{Experimental Result}
	\label{sec:Experimental Results}
	\subsubsection{Materials and Implementation Details.} We validated our proposed Flip Learning method on segmenting the breast nodule in 2D US images. Approved by the local IRB, a total of 1,723 images (with size 448$\times$320)  were collected from 1,129 patients.
	Each image's nodule was annotated with the mask and BBox by sonographers manually (see Fig.~\ref{fig:intro}).
	The dataset was split into 1278, 100 and 345 images for training, validation and independent testing at the patient level with no overlap. The classifier and agents used the same training, validation and testing set. 
	In this study, we implemented our method in $Pytorch$, using an NVIDIA TITAN 2080 GPU. 
	We first trained the classifier with architecture of ResNet18 using AdamW optimiser with learning rate=1e-3 and batch-size=128.	
	Then, the proposed MARL system (ResNet18 backbone) was trained with Adam optimiser in 100 epochs, costing about 1.5 days (learning rate=5e-5 and batch-size=64).
	The superpixels are generated by $OpenCV$ function and indexed from 1 to S. Both agents start from the centre superpixel with index S/2. One agent traverses the superpixels from S/2 to S, the other one traverses reversely from index S/2 to 1. The replay buffer has a size of 8000, and the target $Q$-network copied the parameters of the current $Q$-network every 1200 iterations.
	
	\begin{table}[t]
		\caption{Method comparison (mean$\pm$std). The best WSS results are shown in blue.}\label{compare_different}
		\begin{center}
			\begin{tabular}{c|c|c|c|c|c}
				\toprule
				& \textbf{DICE$\uparrow$} & \textbf{JAC$\uparrow$}& \textbf{CON$\uparrow$} &\textbf{HD$\downarrow$} &\textbf{ASD$\downarrow$}\\
				\hline
				U-net & {93.44$\pm$3.76} & {87.91$\pm$5.02} &{84.76$\pm$9.22} &{15.22$\pm$8.99} & {2.68$\pm$1.79} \\
				\hline
				\hline
				GrabCut & 1.66$\pm$8.91 & 1.07$\pm$5.96 &  \textless10$^{3}$ & 38.11$\pm$124.32&8.11$\pm$27.32\\
				\hline
				Saliency & 43.02$\pm$9.74 & 27.89$\pm$7.93 &-199.90$\pm$189.29 &85.08$\pm$20.04 & 25.86$\pm$4.52 \\
				\hline
				Grad-CAM & 52.66$\pm$14.34 & 38.77$\pm$12.81 &-153.95$\pm$565.80& 61.71$\pm$27.44 & 24.91$\pm$7.08 \\
				\hline
				Grad-CAM++ & 61.99$\pm$11.14 & 45.94$\pm$11.13 & -40.17$\pm$129.12 & 57.11$\pm$24.23 & 22.91$\pm$6.81 \\
				\hline
				SP-RG & 62.66$\pm$14.98 & 47.31$\pm$15.78 & -64.76$\pm$ 519.77& 61.99$\pm$20.60&14.17$\pm$9.22 \\
				\hline
				\hline
				Ours  &\textcolor{blue}{91.12$\pm$2.79} & \textcolor{blue}{83.81$\pm$4.62} &\textcolor{blue}{80.31$\pm$6.90} &\textcolor{blue}{16.95$\pm$7.29}
				&\textcolor{blue}{2.80$\pm$1.33}  \\
				\bottomrule
			\end{tabular}
		\end{center}
	\end{table}
	
	\begin{table*}[t]
		\caption{Ablation study (mean$\pm$std). The best results are shown in blue.}\label{ablation}
		\begin{center}
			\begin{tabular}{c|c|c|c|c|c|c|c}
				\toprule
				\multicolumn{3}{c|}{Strategy} & \multirow{2}*{\textbf{DICE$\uparrow$}} & \multirow{2}* {\textbf{JAC$\uparrow$}} & \multirow{2}* {\textbf{CON$\uparrow$}} &\multirow{2}* {\textbf{HD$\downarrow$}}&\multirow{2}* {\textbf{ASD$\downarrow$}}\\
				
				\cline{1-3}
				
				\textit{MARL} & \textit{IDR} & \textit{C2F} & &&&&\\
				\hline
				
				$\times$ & $\times$ & $\times$, $F$ & 47.95$\pm$19.17& 33.26$\pm$16.59& -314.65$\pm$1150.81 &27.78$\pm$9.28&5.21$\pm$4.78\\
				
				$\times$ & $\times$ & $\times$, $C$ & 77.33$\pm$10.22 &65.09$\pm$12.77&  38.11$\pm$10.88 &21.81$\pm$9.47&3.43$\pm$1.96\\
				\hline
				$\surd$ & $\times$ & $\times$, $F$&46.27$\pm$15.32& 31.99$\pm$16.72& -177.89$\pm$820.77&28.32$\pm$9.87&5.09$\pm$3.71
				\\
				
				$\surd$ & $\times$ & $\times$, $C$&77.64$\pm$12.33& 65.23$\pm$12.08& 37.33$\pm$12.05&22.08$\pm$9.31&3.62$\pm$1.45 \\
				\hline
				$\surd$ & $\surd$ & $\times$, $F$  &
				79.08$\pm$8.77 &69.87$\pm$11.03&41.67$\pm$14.99&21.73$\pm$7.14& 3.68$\pm$1.77\\
				
				$\surd$ & $\surd$ & $\times$, $C$  &
				83.41$\pm$9.52 &72.66$\pm$10.72&53.66$\pm$12.83&20.01$\pm$5.32& 3.21$\pm$1.41\\
				\hline
				$\surd$ & $\times$ & $\surd$ &83.98$\pm$11.27&74.85$\pm$11.22&56.72$\pm$13.20&19.55$\pm$6.05& 3.00$\pm$1.59\\
				\hline
				$\surd$ & $\surd$ & $\surd$
				&\textcolor{blue}{91.12$\pm$2.79} & \textcolor{blue}{83.81$\pm$4.62} &\textcolor{blue}{80.31$\pm$6.90} &\textcolor{blue}{16.95$\pm$7.29}
				&\textcolor{blue}{2.80$\pm$1.33} \\
				\bottomrule
			\end{tabular}
		\end{center}
	\end{table*}
	
	\begin{table*}[h]
		\caption{Impact of Annotation Box Shift (mean$\pm$std).}\label{box_shifting}
		\begin{center}
			\begin{tabular}{c|c|c|c|c|c}
				\toprule
				&\textbf{DICE$\uparrow$} & \textbf{JAC$\uparrow$} & \textbf{CON$\uparrow$} & \textbf{HD$\downarrow$} & \textbf{ASD$\downarrow$}\\
				\hline
				\textit{0-10 pixels} & 89.22$\pm$4.12 & 81.22$\pm$3.22 &72.28$\pm$12.18 & 17.02$\pm$7.87 & 2.88$\pm$1.31\\
				\hline
				\textit{10-20 pixels} & 88.16$\pm$6.23 & 79.56$\pm$5.32 &69.73$\pm$14.11 & 16.53$\pm$7.19&2.92$\pm$2.15 \\
				\hline
				\textit{20-30 pixels} & 86.72$\pm$8.33 &  63.44$\pm$6.33&53.09$\pm$11.28&23.84$\pm$9.88&4.12$\pm$2.13 \\
				\bottomrule
			\end{tabular}
		\end{center}
	\end{table*}

	\subsubsection{Quantitative and Qualitative Analysis.} We evaluated the segmentation performance by five metrics, including indicators of Dice similarity coefficient (DICE-\%), Jaccard index (JAC-\%), Conformity(CON-\%), Hausdorff distance (HD-pixel) and Average surface distance (ASD-pixel). These metrics can provide an objective evaluation between the ground truth (GT) and the prediction.
	As shown in Table~\ref{compare_different}, we reported the quantitative results of Flip Learning (Ours) and six other methods (with the same BBox annotations for labels as ours) including, U-net~\cite{ronneberger2015u}, Grad-CAM~\cite{selvaraju2017grad}, Grad-CAM++~\cite{chattopadhay2018grad}, GrabCut~\cite{rother2004grabcut}, Saliency~\cite{hou2007saliency}, and superpixel-level region growing based on intensity distribution (SP-RG). It can be seen that our methods outperform all the traditional methods and the most common WSS approaches (i.e., Grad-CAM and Grad-CAM++). It is also noted that the results of our proposed method are very closed to that of U-net, which is a \textit{fully-supervised} method.
	
	In the ablation studies, we conducted different experiments to test the superiority of our proposed MARL, IDR, and C2F. 
	CSR is set as the basic reward for all the methods.
	As shown in Table~\ref{ablation}, MARL may not boost the performance in both fine and coarse superpixel-based environment (row 1-4). However, it can save almost 50\% running times compared with the single-agent situation. 
	Interesting to see that interacting with the fine-superpixel-based environment directly (i.e., one-stage) makes agents learn difficultly. Without enough information for guiding agents' action in the huge search space, they may fail to segment accurately. Thus, there will remain many segmentation residuals, which can cause obvious performance degradation (row 1-6). 
	The contribution of IDR and C2F can be observed in the last 4 rows: equipping each of them separately can boost the accuracy, while combining them will obtain a great improvement in all the evaluation metrics.
	To test the sensitivity of our approach to box annotation, we validated it on different box shifting levels, including 1) 0-10 pixels, 2) 10-20 pixels and 3) 20-30 pixels. 
	The results reported in Table~\ref{box_shifting} indicate that our methods can perform well though the box's centre is shifting.
	
\begin{figure*}[!t]
		\centering
		\includegraphics[width=0.92\linewidth]{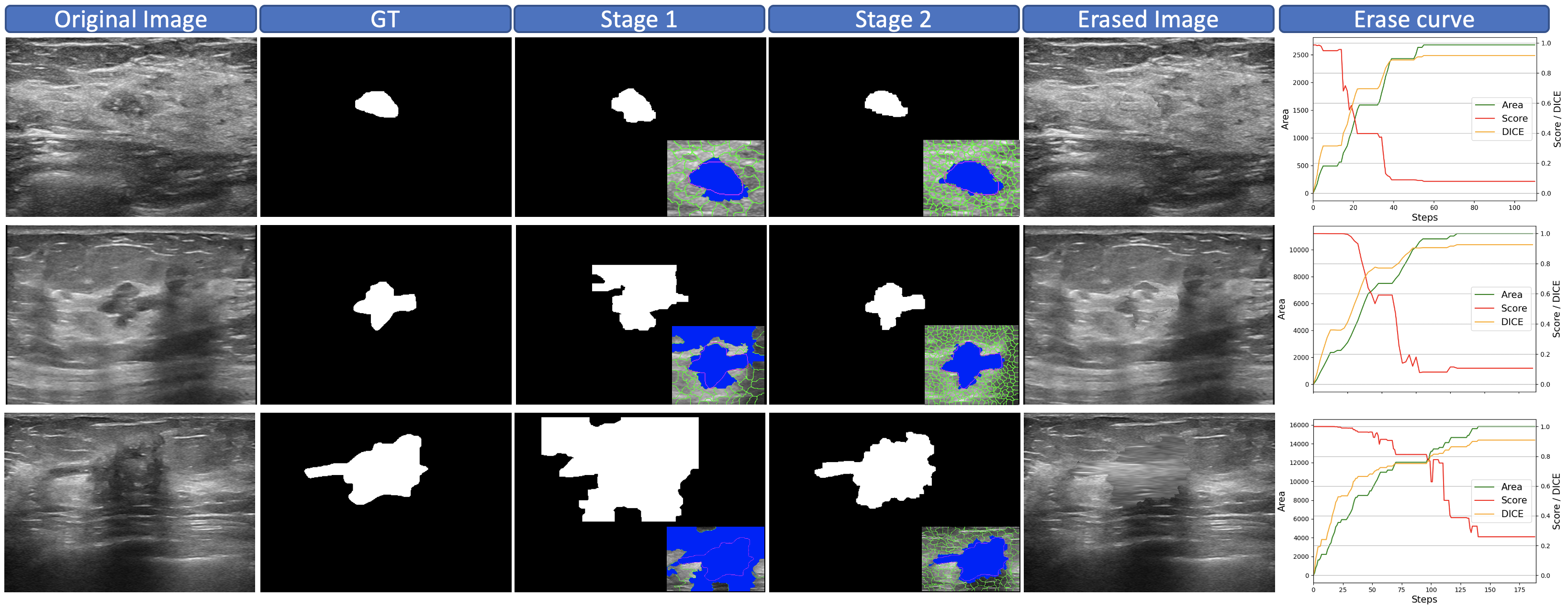}
		\caption{Typical cases of Flip Learning. Mask predictions have been post-processed. Note that the classification tags of original images are `nodule', and after two-stage erasing, their tags will flip to `normal tissue'. The erase curves show the variation of classification scores during erasing.}
		\label{fig:result}
	\end{figure*}

	Fig.~\ref{fig:result} shows three typical cases of our proposed method. Compared with the result of stage one, it can be seen that the final result of stage two obtains a more accurate boundary and overall mask, which is very close to the GT. 
	The erase curves shown in the last column indicate the relationship among erased area size (green), DICE (yellow), and classification score (red). 
	It can be observed that through erasing, the DICE and erased area are gradually increasing, and their variation is nearly synchronous. Moreover, the classification score curve will decrease continuously, and the classification tag will be flipped from `nodule' to `normal tissue', which proves the effectiveness of our Flip Learning approach.
	
	\section{Conclusion}
	\label{sec:conclusion}
	We propose a novel Flip Learning framework for nodule segmentation in 2D US images. We use MARL to erase the nodule from the superpixel-based BBox to flip its classification tag. We develop two rewards, including CSR and IDR, for overcoming the under- and over-segmentation, respectively. Moreover, we propose to adopt a C2F learning strategy in two stages, which can achieve more accurate results than a one-stage method. Experiments on our large in-house dataset validate the efficacy of our method. Our patch-level copy-paste filling strategy is limited in some cases. Thus, in the future, we will explore a more general background filling approach (e.g. GAN), to generate a more accurate background for different types of images.
	
 	\subsubsection{Acknowledgment.}
 	This work was supported by the National Key R\&D Program of China (No. 2019YFC0118300), Shenzhen Peacock Plan (No. KQTD20160\\-53112051497, KQJSCX20180328095606003), Royal Academy of Engineering under the RAEng Chair in Emerging Technologies (CiET1919/19) scheme, EPSRC TUSCA (EP/V04799X/1) and the Royal Society CROSSLINK Exchange Programme (IES/NSFC/201380).

	\bibliographystyle{splncs04}
	\bibliography{paper1545}

\begin{thebibliography}{10}
\providecommand{\url}[1]{\texttt{#1}}
\providecommand{\urlprefix}{URL }
\providecommand{\doi}[1]{https://doi.org/#1}

\bibitem{2012SEEDS}
Bergh, M.V.D., Boix, X., Roig, G., Capitani, B.D., Gool, L.V.: Seeds:
  Superpixels extracted via energy-driven sampling. In: European Conference on
  Computer Vision. Springer (2012)

\bibitem{chattopadhay2018grad}
Chattopadhay, A., Sarkar, A., Howlader, P., Balasubramanian, V.N.: Grad-cam++:
  Generalized gradient-based visual explanations for deep convolutional
  networks. In: 2018 IEEE Winter Conference on Applications of Computer Vision
  (WACV). pp. 839--847. IEEE (2018)

\bibitem{dai2015boxsup}
Dai, J., He, K., Sun, J.: Boxsup: Exploiting bounding boxes to supervise
  convolutional networks for semantic segmentation. In: Proceedings of the IEEE
  international conference on computer vision. pp. 1635--1643. IEEE (2015)

\bibitem{hou2007saliency}
Hou, X., Zhang, L.: Saliency detection: A spectral residual approach. In: 2007
  IEEE Conference on computer vision and pattern recognition. pp.~1--8. IEEE
  (2007)

\bibitem{khoreva2017simple}
Khoreva, A., Benenson, R., Hosang, J., Hein, M., Schiele, B.: Simple does it:
  Weakly supervised instance and semantic segmentation. In: Proceedings of the
  IEEE conference on computer vision and pattern recognition. pp. 876--885.
  IEEE (2017)

\bibitem{kulharia2020box2seg}
Kulharia, V., Chandra, S., Agrawal, A., Torr, P., Tyagi, A.: Box2seg: Attention
  weighted loss and discriminative feature learning for weakly supervised
  segmentation. In: European Conference on Computer Vision. pp. 290--308.
  Springer (2020)

\bibitem{liu2019deep}
Liu, S., Wang, Y., Yang, X., Lei, B., Liu, L., Li, S.X., Ni, D., Wang, T.: Deep
  learning in medical ultrasound analysis: a review. Engineering
  \textbf{5}(2),  261--275 (2019)

\bibitem{minaee2020image}
Minaee, S., Boykov, Y., Porikli, F., Plaza, A., Kehtarnavaz, N., Terzopoulos,
  D.: Image segmentation using deep learning: A survey. arXiv preprint
  arXiv:2001.05566  (2020)

\bibitem{mnih2015human}
Mnih, V., Kavukcuoglu, K., Silver, D., Rusu, A.A., Veness, J., Bellemare, M.G.,
  Graves, A., Riedmiller, M., Fidjeland, A.K., Ostrovski, G., et~al.:
  Human-level control through deep reinforcement learning. Nature
  \textbf{518}(7540),  529--533 (2015)

\bibitem{pont2016multiscale}
Pont-Tuset, J., Arbelaez, P., Barron, J.T., Marques, F., Malik, J.: Multiscale
  combinatorial grouping for image segmentation and object proposal generation.
  IEEE transactions on pattern analysis and machine intelligence
  \textbf{39}(1),  128--140 (2016)

\bibitem{ronneberger2015u}
Ronneberger, O., Fischer, P., Brox, T.: U-net: Convolutional networks for
  biomedical image segmentation. In: International Conference on Medical image
  computing and computer-assisted intervention. pp. 234--241. Springer (2015)

\bibitem{rother2004grabcut}
Rother, C., Kolmogorov, V., Blake, A.: ``grabcut" interactive foreground
  extraction using iterated graph cuts. ACM transactions on graphics (TOG)
  \textbf{23}(3),  309--314 (2004)

\bibitem{schaul2016prioritized}
Schaul, T., Quan, J., Antonoglou, I., Silver, D.: Prioritized experience
  replay. In: ICLR (Poster) (2016)

\bibitem{selvaraju2017grad}
Selvaraju, R.R., Cogswell, M., Das, A., Vedantam, R., Parikh, D., Batra, D.:
  Grad-cam: Visual explanations from deep networks via gradient-based
  localization. In: Proceedings of the IEEE international conference on
  computer vision. pp. 618--626. IEEE (2017)

\bibitem{siegel2021cancer}
Siegel, R.L., Miller, K.D., Fuchs, H.E., Jemal, A.: Cancer statistics, 2021.
  CA: a Cancer Journal for Clinicians  \textbf{71}(1),  7--33 (2021)

\bibitem{tu2008auto}
Tu, Z.: Auto-context and its application to high-level vision tasks. In: 2008
  IEEE Conference on Computer Vision and Pattern Recognition. pp.~1--8. IEEE
  (2008)

\bibitem{van2016deep}
Van~Hasselt, H., Guez, A., Silver, D.: Deep reinforcement learning with double
  q-learning. In: Thirtieth AAAI conference on artificial intelligence (2016)

\bibitem{villani2008optimal}
Villani, C.: Optimal transport: old and new, vol.~338. Springer Science \&
  Business Media (2008)

\bibitem{wei2017object}
Wei, Y., Feng, J., Liang, X., Cheng, M.M., Zhao, Y., Yan, S.: Object region
  mining with adversarial erasing: A simple classification to semantic
  segmentation approach. In: Proceedings of the IEEE conference on computer
  vision and pattern recognition. pp. 1568--1576. IEEE (2017)

\bibitem{yang2021searching}
Yang, X., Huang, Y., Huang, R., Dou, H., Li, R., Qian, J., Huang, X., Shi, W.,
  Chen, C., Zhang, Y., et~al.: Searching collaborative agents for multi-plane
  localization in 3d ultrasound. Medical Image Analysis p. 102119 (2021)

\bibitem{Zhang2017mixup}
Zhang, H., Ciss{\'{e}}, M., Dauphin, Y.N., Lopez{-}Paz, D.: mixup: Beyond
  empirical risk minimization. arXiv preprint arXiv:1710.09412  (2017)

\bibitem{zhou2016learning}
Zhou, B., Khosla, A., Lapedriza, A., Oliva, A., Torralba, A.: Learning deep
  features for discriminative localization. In: Proceedings of the IEEE
  conference on computer vision and pattern recognition. pp. 2921--2929. IEEE
  (2016)

\bibitem{zhou2021deep}
Zhou, S.K., Le, H.N., Luu, K., Nguyen, H.V., Ayache, N.: Deep reinforcement
  learning in medical imaging: A literature review. arXiv preprint
  arXiv:2103.05115  (2021)

\end{thebibliography}
\end{document}